\documentclass[letterpaper, 10 pt, conference]{ieeeconf}  
\IEEEoverridecommandlockouts                              
\usepackage{graphics} 
\usepackage{epsfig} 
\usepackage{times} 
\usepackage{amsmath} 
\usepackage{cite}
\usepackage{tablefootnote}
\usepackage{capt-of}
\usepackage{atbegshi}
\AtBeginDocument{\AtBeginShipoutNext{\AtBeginShipoutDiscard}}

\begin{document}
\title{\LARGE \bf Domain Randomization and Generative Models for Robotic Grasping}%
\author{Josh Tobin$^{1 2}$, Lukas Biewald$^{3 ^{\ast}}$, Rocky Duan$^{4 ^{\ast}}$, Marcin Andrychowicz$^{1}$, Ankur Handa$^{5 ^{\ast}}$, Vikash Kumar$^{2 ^{\ast}}$,%
        \\Bob McGrew$^{1}$, Alex Ray$^{1}$, Jonas Schneider$^{1}$, Peter Welinder$^{1}$, Wojciech Zaremba$^{1 ^{\dagger}}$, Pieter Abbeel$^{2 4 ^{\dagger}}$}%
\thanks{$^{1}$ OpenAI \hspace{15pt} $^{2}$ UC Berkeley \hspace{15pt} $^{3}$ Weights and Biases, Inc}%
\thanks{$^{4}$ Embodied Intelligence \hspace{15pt} $^{5}$ NVIDIA}%
\thanks{$^{\ast}$ Work done while at OpenAI \hspace{15pt} $^{\dagger}$ Equal advising}%
\thanks{Correspondence to {\tt\small josh@openai.com}}%
\makeatletter
\let\@oldmaketitle\@maketitle
\renewcommand{\@maketitle}{\@oldmaketitle
\medskip
\includegraphics[width=1.0\linewidth]{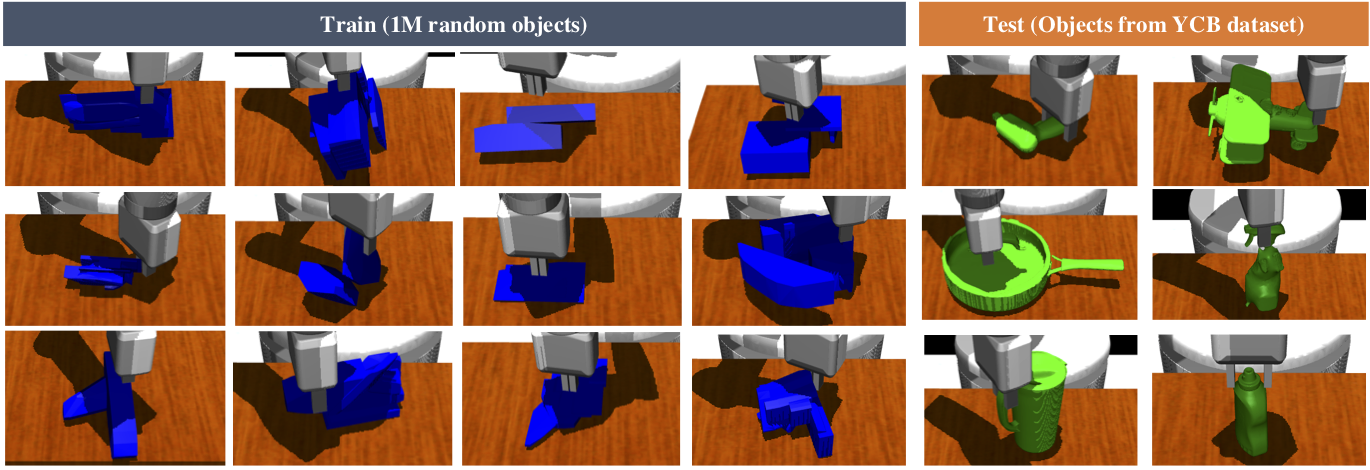}
\captionof{figure}{An overview of our approach. Since creating large numbers of realistic object models is challenging, we train our deep autoregressive model architecture on millions of
         unrealistic procedurally
         generated objects (indicated in blue above) and billions of unique
         grasp attempts. 
         At test time, our model generalizes to realistic objects
         from the YCB dataset (indicated in green above) \cite{calli2015ycb} 
         with 92\% success rate.}\medskip
}
\maketitle
\thispagestyle{empty}
\pagestyle{empty}

\begin{abstract}

Deep learning-based robotic grasping has made significant
progress thanks to algorithmic improvements and increased data availability. 
However, state-of-the-art models are often trained on as few as hundreds or thousands 
of unique object instances, and as a result generalization can be a challenge. 

In this work, we explore a novel data generation pipeline for training a deep 
neural network to perform grasp planning that applies the idea of \emph{domain 
randomization} to object synthesis. We generate millions of unique, unrealistic 
procedurally generated objects, and train a deep neural network to perform grasp 
planning on these objects. 

Since the distribution of successful grasps for a given object can be highly 
multimodal, we propose an autoregressive grasp planning model that maps sensor
inputs of a scene to a probability distribution over possible grasps. This model allows us to sample grasps 
efficiently at test time (or avoid sampling entirely).

We evaluate our model architecture and data generation pipeline in simulation and the real world.
We find we can achieve a $>$90\% success rate on previously unseen realistic objects 
at test time in simulation despite having only been trained on random objects. 
We also demonstrate an 80\% success rate on real-world grasp attempts despite having only been trained on random simulated objects.
\end{abstract}

\section{INTRODUCTION}

Robotic grasping remains one of the core unsolved problems in manipulation. 
The earliest robotic grasping methods used 
analytical knowledge of a scene to compute an optimal grasp for an 
object \cite{murray1994mathematical, bicchi2000robotic, nguyen1988constructing, 
             prattichizzo2016grasping, rodriguez2012caging, schulman2011grasping}. 
Assuming a contact model and a heuristic for the
likelihood of success of a grasp, analytical methods can provide guarantees 
about grasp quality, but they often fail
in the real world due to inconsistencies in the simplified object and contact 
models, the need for accurate 3D models of the objects in question, 
and sensor inaccuracies \cite{bohg2014data}.

As a result, significant research attention has been given to data-driven 
grasp synthesis methods \cite{bohg2014data, miller2003automatic, pelossof2004svm, 
                              goldfeder2007grasp, morales2004using,
                              montesano2008learning, saxena2008robotic}. 
These algorithms avoid some of the challenges of 
analytic methods by sampling potential grasps and ranking them according to a 
learned function that maps sensor inputs to an estimate of a chosen heuristic.

Recently, several works have explored using deep neural networks to approximate 
the grasp heuristic function \cite{levine2016learning, pinto2016supervision, 
mahler2017dex, johns2016deep}. The promise of deep neural networks 
for learning grasp heuristics is that with diverse training data, 
deep models can learn features that deal with the edge cases that 
make real-world grasping challenging. 

A core challenge for deep learning grasp quality heuristics is data availability. 
Due to the difficulty and
expense of collecting real-world data and due to 
the limited 
availability of high-quality 3D object meshes, current approaches use 
as few as hundreds or thousands of unique object instances, 
which may limit generalization. 
In contrast, ImageNet \cite{krizhevsky2012imagenet}, the standard benchmark for image
classification, has about 15M unique images from 22K categories.

In order to increase the availability of training data in simulation, we explore 
applying the idea of \emph{domain randomization} \cite{sadeghi2016cad, tobin2017domain} 
to the creation of 3D object meshes. Domain randomization 
is a technique for learning models that work in a test domain after only 
training on low-fidelity simulated data by randomizing all non-essential aspects 
of the simulator. One of the core hypotheses of this work is that by training
on a wide enough variety of unrealistic procedurally generated object meshes,
our learned models will generalize to realistic objects. 

Previous work in deep learning for grasping has focused on learning a function 
that estimates the quality of a given grasp given observations of the scene. 
Choosing grasps on which to perform this estimate has received comparatively little 
attention. Grasps are typically
chosen using random sampling or by solving a small optimization problem online. 
The second goal of this paper is to propose a deep learning-based method for 
choosing grasps to evaluate. Our hypothesis is that a learned model for grasp
sampling will be more likely to find high-quality grasps for challenging objects
and will do so more efficiently. 

We use an autoregressive model architecture \cite{larochelle2011neural, 
oord2016pixel, oord2016wavenet} that maps sensor inputs to a 
probability distribution over grasps that corresponds to the model's weighted
estimate of the likelihood of success of each grasp. After training, highest 
probability grasp according to the distribution succeeds on 89\% of test objects and the 20 highest 
probability grasps contain a successful grasp for 
96\% of test objects. 
In order to determine which grasp to execute on the 
robot, we collect a second observation in the form of an image from the 
robot's hand camera and train a second model to choose the most promising grasp
among those sampled from the autoregressive model, resulting in a success rate of 92\%.

The contributions of this paper can be summarized as follows:
\begin{itemize}
\item We explore the effect of training a model for grasping using unrealistic 
      procedurally generated objects and show that such a model can 
      achieve similar success to one trained on a realistic object distribution.
      (Another paper \cite{bousmalis2017using} developed concurrently to this 
      one explored a similar idea and reached similar conclusions.)
\item We propose a novel generative model architecture and training methodology 
      for learning a sampling distribution for grasps to evaluate.
\item We evaluate our object generation, training, and
      sampling algorithms in simulated scenes and find that we can achieve 
      an 84\% success rate on random objects and 92\% 
      success rate on previously unseen real-world objects despite 
      training only on non-realistic randomly generated objects. 
\item We demonstrate that we can deploy these models in real-world grasping experiments with an 80\% success rate despite having been trained entirely in simulation.
\end{itemize}

\section{RELATED WORK}

\subsection{Domain Randomization}
Domain randomization involves randomizing non-essential aspects of the training 
distribution in order to better generalize to a difficult-to-model test 
distribution. This idea has been employed in robotics since at least
1997, when Jakobi proposed the ``Radical Envelope of Noise Hypothesis", the idea
that evolved controllers can be made more robust by completely randomizing all
aspects of the simulator that do not have a basis in reality and slightly randomizing
all aspects of the simulator that do have a basis in reality
\cite{jakobi1997evolutionary}. 
Recently domain randomization has shown promise in transferring 
deep neural networks for robotics tasks from simulation to the real world by 
randomizing physics \cite{mordatch2015ensemble} and appearance 
properties \cite{sadeghi2016cad, tobin2017domain, zhang2017sim}. 

In another work developed concurrently with this one \cite{bousmalis2017using}, the authors 
reach a similar conclusion about the utility of procedurally generated objects 
for the purpose of robotic grasping. In contrast to this work, theirs focuses 
on how to combine simulated data with real grasping data to achieve successful 
transfer to the real world, but does not focus on achieving a high overall 
success rate. Our paper instead focuses on how to achieve the best possible 
generalization to novel objects. Ours also has a comparable real-world success rate despite not using any real-world training data. 

\subsection{Autoregressive models}
This paper uses an autoregressive architecture to model a distribution 
over grasps conditioned on observations of an object. 
Autoregressive models leverage the fact that an $N$-dimensional probability 
distribution $p(X)$ can be factored as $\prod_{n=1}^N p(x_n \mid x_1, \cdots, 
x_{n-1})$ for any choice of ordering of $1$-dimensional variables $x_i$. The 
task of modeling the distribution then consists of modeling each 
$p(x_n \mid x_1, \cdots, x_{n-1})$ \cite{larochelle2011neural}. In contrast to 
Generative Adversarial Networks \cite{NIPS2014_5423}, another popular form for a
deep generative model, autoregressive models can directly compute the likelihood 
of samples, which is advantageous for tasks like grasping in which finding the 
highest likelihood samples is important. Autoregressive models have been used 
for density estimation and generative modeling in image domains 
\cite{larochelle2011neural, gregor2013deep, germain2015made}
and have been
shown to perform favorably on challenging image datasets like ImageNet 
\cite{oord2016pixel, NIPS2016_6527}. Autoregressive models have also 
been successfully applied to other forms of data including in topic modeling 
\cite{larochelle2012neural} and audio generation \cite{oord2016wavenet}. 

\subsection{Robotic grasping}
Grasp planning methods fall into one of two categories: \emph{analytical} methods
and \emph{empirical} methods \cite{sahbani2012overview}. 

Analytical methods use a contact model and knowledge of an object's 3D shape to find grasps that maximize 
a chosen metric like the ability of the grasp to resist external wrenches \cite{prattichizzo2016grasping} or constrain
the object's motion \cite{rodriguez2012caging}. Some methods attempt to make these estimates
more robust to gripper and object pose uncertainty and sensor error by instead 
maximizing the expected value of a metric under uncertainty \cite{kehoe2013cloud, weisz2012pose}.

Most approaches use simplified Coulomb friction and rigid body 
modeling for computational tractability \cite{murray1994mathematical, prattichizzo2016grasping}, but some have explored
more realistic object and friction models \cite{bicchi2000robotic, prattichizzo2012manipulability, rosales2012synthesis}.
Typically, grasps for an object are selected based on sensor data by registering images to a known database of 3D models 
using a traditional computer vision pipeline \cite{ciocarlie2014towards,  hinterstoisser2011multimodal, xie2013multimodal}
or a deep neural network \cite{gupta2015aligning, zeng2017multi}. 

Empirical methods instead attempt to maximize the value of a quality metric through
sampling \cite{sahbani2012overview}. Many approaches use simulation to 
evaluate classical grasp quality metrics \cite{miller2003automatic, pelossof2004svm, goldfeder2007grasp, miller2004graspit}.
Others use human labeling or self-supervised learning to measure success 
\cite{morales2004using,  montesano2008learning, saxena2008robotic}. 

Most techniques estimate the value of the quality metric in real-world trials
using traditional computer vision and learning techniques \cite{saxena2008learning, le2010learning, saxena2008robotic, pelossof2004svm}.
More recently, deep learning has been employed to learn a mapping directly from
sensor observations to grasp quality or motor torques.

\subsection{Deep learning for robotic grasping}

Work in deep learning for grasping can be categorized by how training data is collected 
and how the model transforms noisy observations into grasp candidates. 

Some approaches use hand-annotated real-world grasping trials to provide training 
labels \cite{lenz2015deep}. However, hand-labeling is challenging to scale to 
large datasets. To alleviate this problem, some work explores automated 
large-scale data collection \cite{levine2016learning, pinto2016supervision}. 
Others have explored replacing real data with synthetic depth data at training 
time \cite{mahler2017dex, mahler2017dex3, viereck2017learning, johns2016deep}, 
or combining synthetic RGB images with real images 
\cite{bousmalis2017using}. In many cases, simulated data appears to be effective
in replacing or supplementing real-world data in robotic grasping. Unlike our 
approach, previous work using synthetic data uses small datasets of up to a 
few thousand of realistic object meshes. 

One commonly used method for sampling grasps is to learn a visuomotor control 
policy for the robot that allows it to iteratively refine its grasp target as 
it takes steps in the environment. Levine and co-authors 
learn a prediction network $g(I_t, v_t)$ that takes an observation $I_t$ and 
motor command $v_t$ and outputs a predicted probability of a successful grasp 
if $v_t$ is executed \cite{levine2016learning}. The cross-entropy method is used to greedily select the $v_t$ 
that maximizes $g$. Viereck and co-authors instead learn a 
function $d(I_t, v_t)$ that maps the current observation and an action to an estimate 
of the distance to the nearest successful grasp after performing $v_t$ \cite{viereck2017learning}. Directions 
are sampled and a constant step size is taken in the direction with the minimum 
value for $d$. In contrast to visuomotor control strategies, planning approaches 
like ours avoid the local optima of greedy execution. 

Another strategy to choose a grasp using a deep learning is to sample 
grasps and score them using a deep neural network 
of the form $f(I, \boldsymbol{g}) \rightarrow s$, where $I$ are the observation(s) of the 
scene, $\boldsymbol{g}$ is a selected grasp, and $s$ is the score for the selected grasp 
\cite{mahler2017dex, mahler2017dex3, johns2016deep, yan2017learning}. 
These techniques differ in terms of how they
sample grasps to evaluate at test time. Most commonly they directly optimize 
$g$ using the cross-entropy method \cite{mahler2017dex, mahler2017dex3, yan2017learning}. 
In contrast to these approaches, our approach 
jointly learns a grasp scoring function and a sampling distribution, allowing for
efficient sampling and avoiding exploitation by the optimization 
procedure of under- or over-fit regions of the grasp score function. 

Other approaches take a multi-step approach, starting with a coarse 
representation of the possible grasps for an object and then exhaustively 
searching using a learned heuristic \cite{lenz2015deep} or modeling the score 
function jointly for all possible coarse grasps \cite{johns2016deep}. Once
a coarse grasp is sampled, it is then fine-tuned using a separate network 
\cite{lenz2015deep} or interpolation \cite{johns2016deep}. By using an 
autoregressive model architecture, we are able to directly learn a high-
dimensional ($20^4$ or $20^6$-dimensional) multimodal probability distribution.

\section{METHOD}
Our goal is to learn a mapping that takes one or more 
observations $I = \{I_j\}$ of a scene and outputs a grasp $\boldsymbol{g}$ to attempt 
in the scene. The remainder of the section describes the data generation pipeline,
model architecture, and training procedure used in our method.
\subsection{Data collection}
\begin{figure}[h!]
    \centering
    \includegraphics[width=1.0\linewidth]{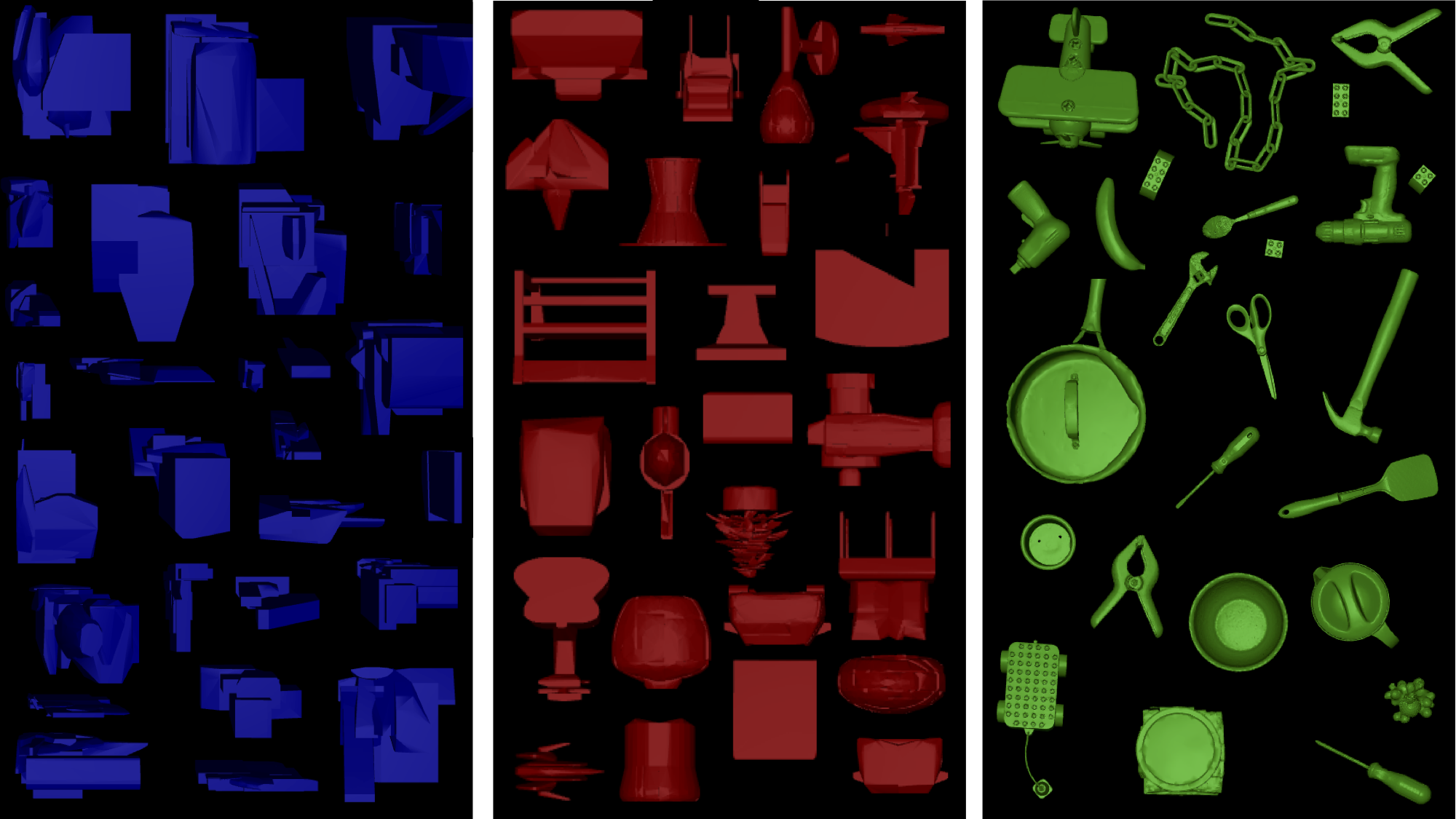}
    \caption{Examples of objects used in our experiments.
    Left: procedurally generated random objects. Middle: objects from the ShapeNet object dataset. Right: objects from the YCB object dataset.}
    \label{}
\end{figure}
We will describe the process of generating training objects, and then 
the process of sampling grasps for those objects.

\subsubsection{Object generation}
One of our core hypotheses is that training on a diverse array 
of procedurally generated objects can produce comparable performance to training 
on realistic object meshes. Our procedurally generated objects 
were formed as follows:
\begin{enumerate}
\item Sample a random number $n_p \in \{1, \cdots 15\}$
\item Sample $n_p$ primitive meshes from our object primitive dataset
\item Randomly scale each primitive so that all dimensions are between 1 and 
15cm 
\item Place the meshes sequentially so that each mesh intersects with at least one of 
the preceding meshes
\item Rescale the final object to approximate the size distribution observed 
in our real object dataset 
\end{enumerate}

To build a diverse object primitive dataset, we took the more than 40,000 
object meshes found in the ShapeNet object dataset \cite{chang2015shapenet} 
and decomposed them into more than 400,000 convex parts using 
V-HACD\footnote{https://github.com/kmammou/v-hacd}. Each 
primitive is one convex part. 

We compared this object generation procedure against a baseline of training 
using rescaled ShapeNet objects.

\subsubsection{Grasp sampling and evaluation}
We sample grasps uniformly at random from a discretized $4$- or $6$-dimensional 
grasp space ($4$-dimensional when attention is restricted to upright grasps) 
corresponding to the $(x, y, z)$ coordinates of the center of the gripper and 
an orientation of the gripper about that point. We discretize each dimension 
into $20$ buckets. The buckets are the relative location of the 
grasp point within the bounding box of the object -- e.g., an $x$-value of $0$ 
corresponds to a grasp at the far left side of the object's bounding box and 
a coordinate of $19$ corresponds to a grasp at the far right. 

Grasps that penetrate the object or for which the gripper would not 
contact the object when closed can be instantly rejected. The remainder are 
evaluated in a physics simulator. 

For each grasp attempt, we also collect a depth image from the robot's hand 
camera during the approach to be used to train the grasp evaluation function.

\subsection{Model architecture}

\begin{figure}[h!]
    \centering
    \includegraphics[width=1.0\linewidth]{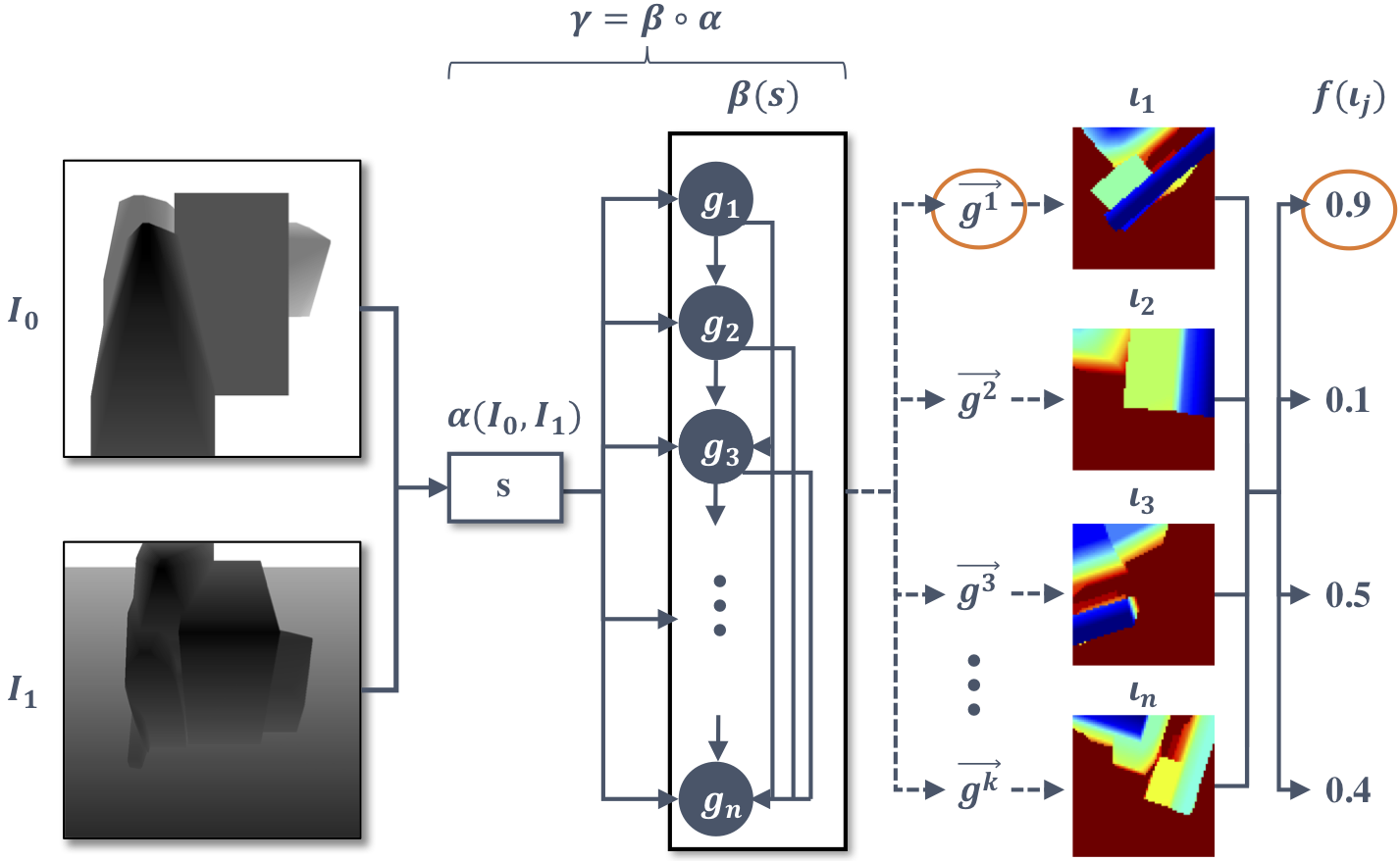}
    \caption{An overview of sampling from our model architecture. Solid lines represent neural networks, and dotted lines represent sampling operations. 
    The model takes as input one or more observations of the target object in the form of depth images. The images are passed to an image representation 
    module $\alpha$, which maps the images to an embedding $s$. The embedding $s$ is the input for the autoregressive module $\beta$, which outputs a distribution over possible grasps $\boldsymbol{g}$ for the object by modeling each dimension $\boldsymbol{g}_i$ of the grasp conditioned on the previous dimensions. We sample $k$ 
    high-likelihood grasps $\vec{g}^1, \cdots \vec{g}^k$ from the model using a beam search. For each of those 
    grasps, a second observation is captured that corresponds to an aligned image 
    in the plane of the potential grasp. A grasp scoring model $f$ maps each 
    aligned image to a score. The grasp with the highest score is selected for execution on the robot.}
    \label{fig:model_architecture}
\end{figure}
The model architecture used for our experiments is outlined in Figure 
\ref{fig:model_architecture}. The model consists of two separate neural 
networks -- a grasp planning module $\gamma(I) = \beta \circ \alpha(I)$ and a 
grasp evaluation model $f$. The grasp planning module is used to sample grasps 
that are likely to be successful. The grasp evaluation model takes advantage of 
more detailed data in the form of a close-up image from a camera on the gripper
of the robot to form a more accurate estimate of the likelihood of each sampled 
grasp to be successful.

The image representation $s = \alpha(I)$ is formed by passing each image through
a separate convolutional neural network. The flattened output of these 
convolutional layers are stacked and passed through several dense layers 
to produce $s$.

The neural network $\beta(s)$ models a probability distribution 
$p_{\beta}(\boldsymbol{g} | s)$ over possible grasps for the object that corresponds to
the normalized probability of success of each grasp. The model $\beta$ 
consists of $n$ submodules $\beta_i$ where $n$ is the dimensionality of the 
grasp. For any grasp $\boldsymbol{g}$, $\beta$ and $\{\beta_i\}$ are related by
$$
\beta(s)(\boldsymbol{g}) = \prod_{i=1}^n \beta_i(g_1, \cdots, g_{i-1}, s),
$$
where $g_i$ are the dimensions of $\boldsymbol{g}$. 

Each $\beta_i$ is a small neural network taking
$s$ and $g_1, \cdots, g_{i-1}$ as input and outputting a softmax over 
the 20 possible values for $g_i$, the next dimension of $g$. We found that sharing weights between the 
$\beta_i$ hurts performance at convergence. 

The grasp evaluation model $f$ takes as input a single observation from the 
hand camera of the robot and outputs a single scalar value corresponding to the 
likelihood of success of that grasp. The model $f$ is parameterized by a 
convolutional neural network with sigmoid output.

\subsection{Training methodology}
Since our method involves capturing depth images from the hand of the robot 
corresponding to samples from $\gamma (I) = \beta \circ \alpha(I)$, 
the entire evaluation procedure 
is not differentiable and we cannot train the model end-to-end using supervised 
learning. As a result, our training procedure involves independently training 
$\gamma$ and $f$. 

Given datasets of objects $D = \{D_1, \cdots D_d\}$, observations 
$I = \{I^1, \cdots I^d\}$ and successful grasps 
$G = \{\boldsymbol{g}_1^1, \cdots \boldsymbol{g}_{m_1}^1, \cdots \boldsymbol{g}_1^2, \cdots \boldsymbol{g}_{m_d}^d\}$, 
$\gamma$ can be optimized by minimizing the negative log-likelihood of $G$ conditioned on the observations $I$ with respect to the parameters $\theta$ of $\gamma$, 
which is given by:
$$
J(\theta) = -\frac{1}{d}\sum_{i=1}^d \frac{1}{m_i} \sum_{j=1}^{m_i} \log p_\theta(\boldsymbol{g}_j^i \mid I^i). 
$$
This can be decomposed as 
$$
-\frac{1}{d}\sum_{i=1}^d \frac{1}{m_i} \sum_{j=1}^{m_i} \sum_{k=1}^n \log \beta_k \left(\alpha(I^i)\right) \left((\boldsymbol{g}_l^i)_{l< k}\right).
$$
This function can be optimized using standard backpropogation and minibatch SGD techniques. \cite{larochelle2011neural}

In practice, $\alpha$ is usually a larger model than $\beta$ and there are often 
tens or hundreds of successful grasp attempts $\{\boldsymbol{g}^i_j\}$ for a single observation  $I^i$. Therefore it is computationally advantageous to perform the 
forward pass and gradient calculations for each $\alpha(I^i)$ once for all 
$\{\boldsymbol{g}^i_j\}$. 

This can be achieved in standard neural network and 
backpropagation libraries by stacking all grasps for a given object so that SGD minibatches consist of pairs $(I_i, \boldsymbol{g}^i_{\{j\}})$ where $\boldsymbol{g}^i_{\{j\}}$ is 
the $m_i \times n$ matrix consisting of all successful grasps for object $i$. 
To deal with differing values for $m_i$, we choose $m = \max(\{m_i\})$ and form 
an $m \times n$ matrix by padding the $m_i \times n$ matrix with arbitrary values. 
We can then write the 
gradient $\nabla_{\theta} J(\theta)$ of our objective function as follows:
$$
-\frac{1}{d}\sum_{i=1}^d \frac{1}{m_i} \nabla_{\theta}\alpha(I^i) \sum_{j=1}^{m}  1_j \sum_{k=1}^n \nabla_{\theta} \log \beta_k \left(\alpha(I^i)\right) \left((\boldsymbol{g}^i_l)_{l < k}\right)
$$
where $1_j$ is an indicator function corresponding to whether the entry in $j$ 
was one of the $m_i$ successful grasps. 

This form of the gradient allows us to compute $\nabla_{\theta} \alpha(I^i)$ once for each $I^i$, which we found to increase training speed by more than a factor of 10 in our experiments.

The grasp evaluation function $f$ is trained using supervised learning. 
Inputs are the hand camera images collected during the data collection process
and labels are an indicator function corresponding to whether the grasp was successful.

\section{EXPERIMENTS}

The goal of our experiments is to answer the following questions:
\begin{enumerate}
    \item Can grasping models trained on unrealistic randomly generated objects perform as well on novel realistic objects as those trained on realistic objects?
    \item How efficiently can we sample grasps from our proposed autoregressive model architecture?
    \item How important is using a large number of unique objects for training grasping models?
    \item How well do models trained using our methodology work in the real world?
\end{enumerate}

\subsection{Experimental setup}
We evaluated our approach by training grasping models on three datasets: 
\begin{itemize}
    \item ShapeNet-1M, a dataset of 1 Million scenes with a single object
     from the ShapeNet dataset with randomized orientation and object scale. 
    \item Random-1M, a dataset of 1 Million scenes with a single object generated 
          at random using the procedure above.
    \item ShapeNet-Random-1M, a dataset with 500,000 scenes from each of the 
          previous datasets.
\end{itemize}
For each object set, we recorded 2,000 grasp attempts per 
object. This number of grasps was selected so that more than 95\% of objects 
sampled had at least one successful grasp to avoid biasing the dataset to 
easier objects\footnote{This number is not closer to 100\% because a small 
percentage of the random objects in our training set are un-graspable with the 
Fetch gripper. }. For data generation, we used a disembodied Fetch gripper to 
improve execution speed. 

We trained the models using the Adam optimizer \cite{kingma2014adam} with a 
learning rate of $10^{-4}$. We trained each model using three random seeds, and 
report the average of the three seeds unless otherwise noted. 

We evaluated the model on $300$ training and hold-out scenes from  ShapeNet-1M and Random-1M, as well as $300$ scenes generated from the $75$ YCB objects with meshes capable of being graseped by our robot's gripper. 

All executions were done using a Fetch robot in the MuJoCo physics simulator \cite{todorov2012mujoco}. 
When evaluating the model, we sampled $k=20$ likely grasps from the model using 
a beam search with beam width $20$. Among these grasps, only the one with the highest score according to $f$ is attempted on the robot. An attempt is considered successful if the robot is able to use this grasp to lift the object by 30cm. 

\subsection{Performance using randomly generated training data}

Figure \ref{fig:perf_table} describes the overall success rate of the algorithm on previously 
seen and unseen data. The full version of our algorithm is able to achieve greater than 90\% success rate on 
previously unseen YCB objects even when training entirely on randomly generated 
objects. Training on 1M random objects performs comparably to training on 1M instances
of realistic objects. 

\begin{figure}[h!]
\centering
\scriptsize
\begin{tabular}{lrrrrr}
\hline
              & ShapeNet & ShapeNet & Random & Random &     \\
 Training set & Train    & Test     & Train  & Test   & Ycb \\
\hline
 ShapeNet-1M & 0.91 & \bf{0.91} & 0.72 & 0.71 & \bf{0.93} \\
 Random-1M   & 0.91 & 0.89 & \bf{0.86} & \bf{0.84} & 0.92 \\
 ShapeNet-Random-1M   & \bf{0.92} & 0.90 & 0.84 & 0.81 & 0.92 \\
\hline
\end{tabular}
\caption{Performance of the algorithm on different synthetic test sets. The 
full algorithm is able to achieve at least 90\% success on previously unseen 
objects from the YCB dataset when trained on any of the three training sets. }
\label{fig:perf_table}
\end{figure}

Note that these results were achieved by limiting the robot to grasps 
in which the gripper is upright. The success rate is around 10\% lower across experiments when using full 6-dimensional grasps. Further experimentation could 
look into whether significantly scaling the amount of training data or using 
a combination of the 4-dimensional training data and 6-dimensional training data 
could improve performance.

Figure \ref{fig:perf_comp} compares the success rate of our 
algorithm to several baselines. In particular, our full method performs 
significantly better than sampling the highest likelihood grasp from the 
autoregressive model alone.

\begin{figure}[h!]
\centering
\scriptsize
\begin{tabular}{lrrrrr}
\hline
                     & ShapeNet & ShapeNet & Random & Random \\
 Training set        & Train &   Test &   Train &   Test &      Ycb \\
\hline
 Full Algorithm     & \bf{0.91} & \bf{0.89} & \bf{0.86} & \bf{0.84} & \bf{0.92} \\ 
 Autoregressive-Only     & 0.89 & 0.86 & 0.80 & 0.76 & 0.89 \\
 Random             & 0.22 & 0.21 & 0.10 & 0.11 & 0.26 \\
 Centroid           & 0.30 & 0.25 & 0.10 & 0.12 & 0.54 \\
\hline
\end{tabular}
\caption{Performance of the algorithm compared to baseline approaches. The 
         Full Algorithm and Autoregressive-Only numbers reported are using 
         models trained on random data. The 
         Autoregressive-Only baseline uses the model $\gamma$ to sample a 
         single high-likelihood grasp, and executes that grasp directly without
         evaluating it with the model $f$. The Random baseline samples a random grasp. 
         The centroid baseline deterministically attempts to grasps the center
         of mass of the object, with the approach angle sampled randomly.}
\label{fig:perf_comp}
\end{figure}

\begin{figure}
    \centering
    \includegraphics[width=0.75\linewidth]{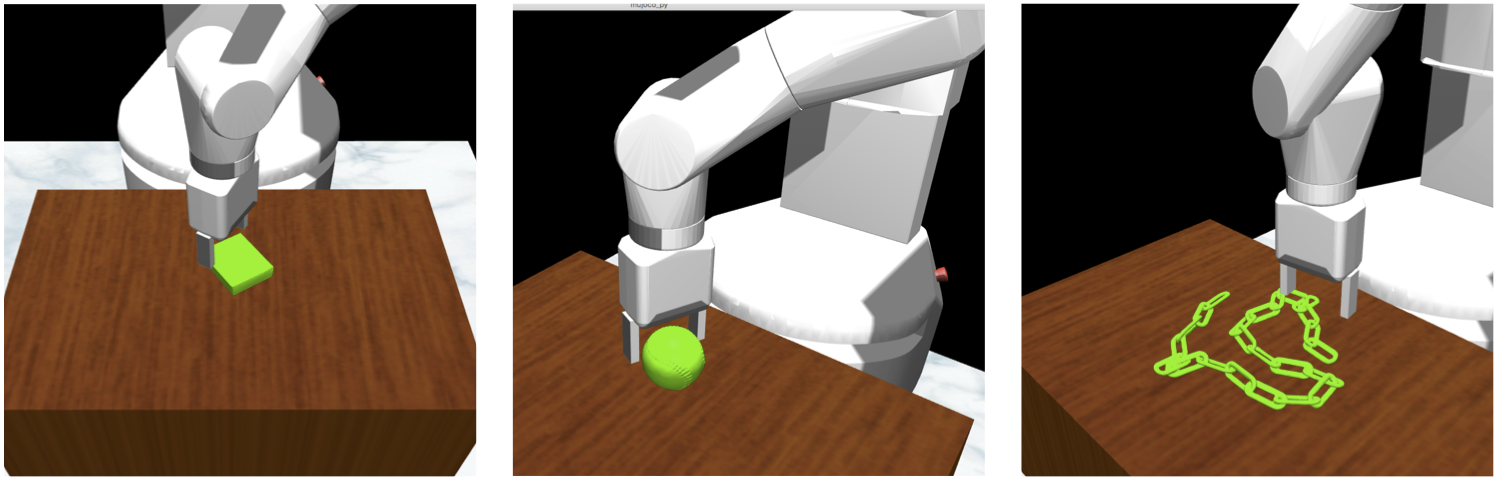}
    \caption{Examples of three observed failure cases for our learned models. Left: the model chooses a grasp for an object that is close to the maximum size of the gripper that collides with the object. Middle: the model chooses a grasp for a large, highly curved object that causes it to slip off. Right: the model fails to find a promising grasp for a highly irregular object. }
    \label{fig:my_label}
\end{figure}
We observed the following three main failure cases for the learned model:
\begin{enumerate}
    \item For objects that are close to the maximum size graspable by the gripper (10cm), the grasp chosen sometimes collides with the object. 
    \item In some cases, the model chooses to grasp a curved object at a narrower point to avoid wider points that may cause collision, causing the gripper to slip off.
    \item The model cannot find reasonable grasp candidates for some highly irregular objects like a chain found in the YCB dataset. 
\end{enumerate}
The learned models primarily failed on objects for which all features are close in size to the maximum allowed by the gripper.  Supplementing the training set with additional objects on the edge of graspability or combining planning with visual servoing could alleviate these cases.

The model also failed for a small number of highly irregular objects like a chain present in the YCB dataset. These failure cases present a larger challenge for the use of random objects in grasping, but additional diversity in the random generation pipeline may mitigate the issue.

\subsection{Efficiency of the autoregressive model}
\begin{figure}[h!]
    \centering
    \includegraphics[width=1.0\linewidth]{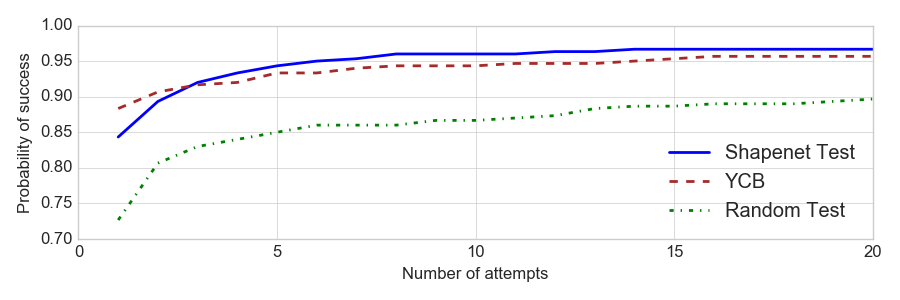}
    \caption{Percentage of objects that a trained model can grasp successfully 
             as a function of number of the number of samples from the 
             autoregressive model attempted. Here, we sample the 20 grasps 
             from the autoregressive model that have the highest likelihood
             according to a beam search and count the number of times success
             occurred on the $n$th attempt for $n < 20$.}
    \label{fig:all_attempts}
\end{figure}

To test how efficiently we are able to sample grasps from our autoregressive model, we looked at the percentage of objects for which the top $k$ most likely grasps according to $\gamma$ contain at least one successful grasp.
Figure \ref{fig:all_attempts} shows that the most likely grasp according to the model succeeds close to 90\% of the time on YCB objects, and the incremental likelihood of choosing a valid grasp saturates between $10$ and $20$ samples, motivating our choice of $20$ for $k$. Note that more objects have successful grasps among the 20 sampled than achieve success 
using our method, suggesting the performance of the grasp evaluator 
$f$ could be a bottleneck to the overall performance of the algorithm.

\subsection{Effect of amount of training data}
\begin{figure}[h!]
    \centering
    \includegraphics[width=1.0\linewidth]{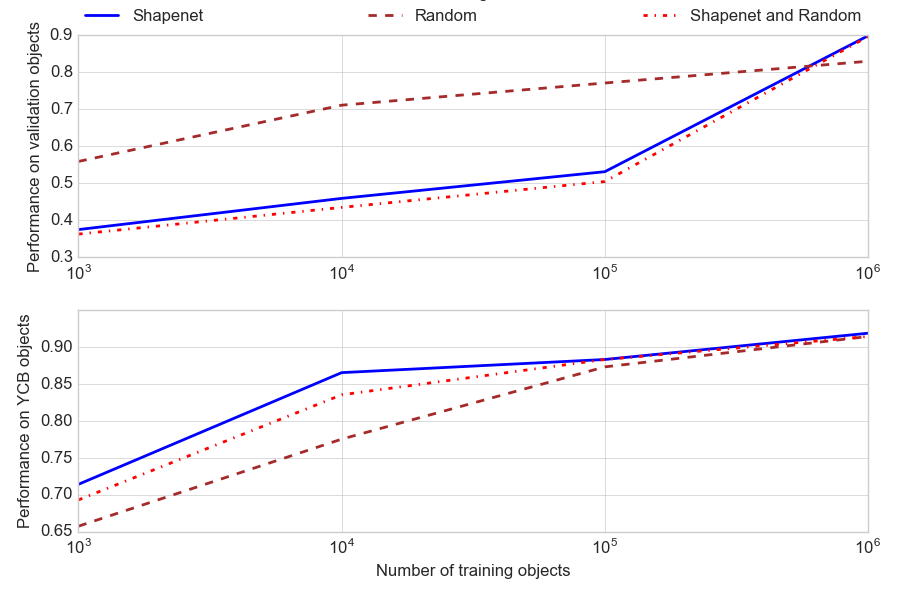}
    \caption{Impact of number of unique training objects used on the performance 
    of the learned model on held out data at test time. Each line represents
    a different training set. The top chart indicates the performance of the 
    models on held out data from the same distribution (i.e., 
    held out ShapeNet or Random objects depending on the training set), 
    and the bottom chart shows performance on out-of-sample objects from the 
    ShapeNet dataset.}
    \label{fig:scaling}
\end{figure}

Figure \ref{fig:scaling} shows the impact of the number of unique objects in 
the training set on the performance of our models in validation data held out 
from the same distribution and out-of-sample test data from the YCB dataset. 
Although with enough data the model trained entirely using randomly generated 
data performs as well as the models trained using realistic data, with smaller 
training sets the more realistic object distributions perform significantly 
better on the test set than does the unrealistic random object distribution. 

Note that performance does not appear to have saturated yet in these examples. 
We conjecture that more training data and more training data diversity 
could help reduce the effects of the first two failure cases above, but may
 not allow the model to overcome the third failure case. 

\subsection{Physical robot experiments}
\begin{figure}[h!]
    \centering
    \includegraphics[width=0.6\linewidth]{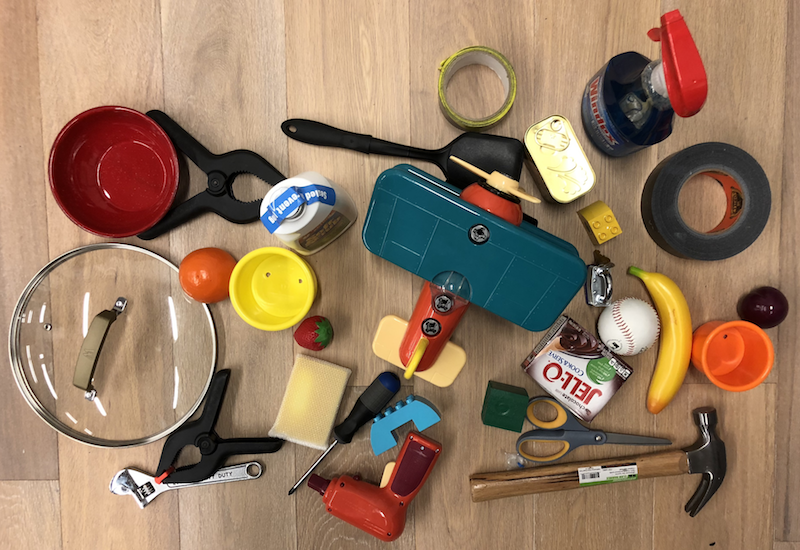}
    \caption{The objects used in our real-world experiments.}
    \label{fig:realims}
\end{figure}

\begin{figure}[h!]
    \centering
    \includegraphics[width=1.0\linewidth]{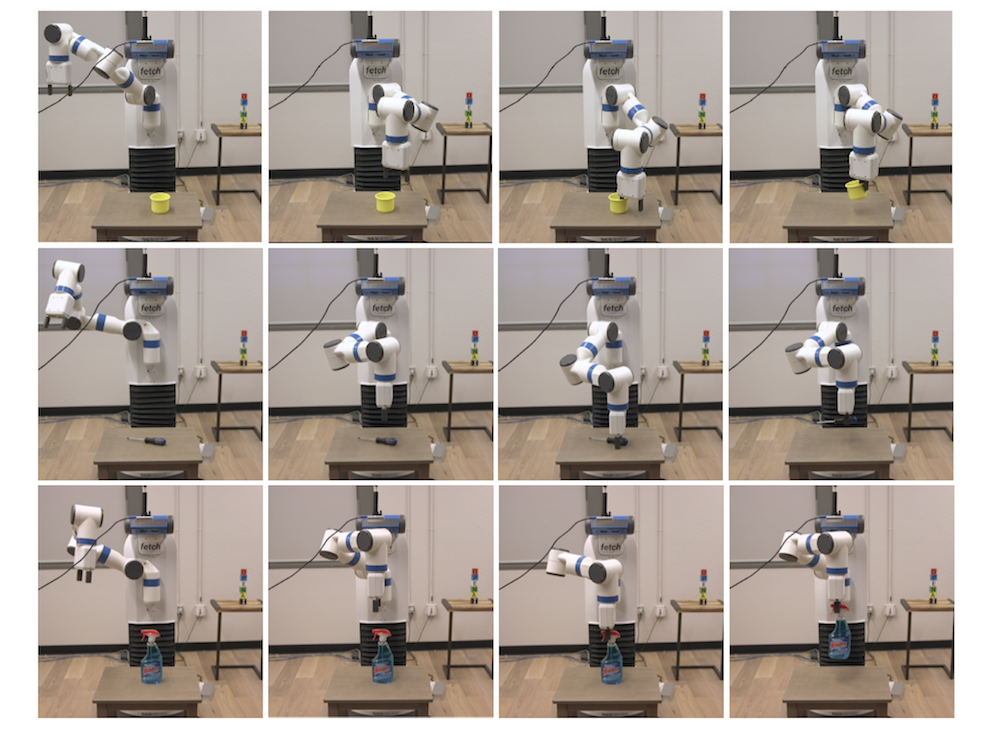}
    \caption{Typical grasp executions on the physical robot.}
    \label{fig:execution}
\end{figure}

We evaluated the ability of models learned on synthetic data using our method to transfer to the real world by attempting to grasp objects from the YCB dataset with a Fetch robot. Figure \ref{fig:realims} shows the objects used in our experiments. Models executed in the real-world were the Autoregressive-Only variant of our method trained on the Random-1M dataset with a single depth image from directly above the scene as input. Figure \ref{fig:execution} depicts typical grasp executions on the Fetch robot during our experiments.

At test time, the depth input was produced by an Intel RealSense D435 \cite{keselman2017intel}. To model the noise in real-world depth images, we followed the method of Mahler et al. \cite{mahler2017dex}, who propose the observation model $I = \alpha \hat{I} + \epsilon$, where $\hat{I}$ is the rendered depth image, $\alpha$ is a Gamma random variable and $\epsilon$ is a zero-mean Gaussian random variable.

We tested the learned model on 30 previously unseen objects chosen to capture the diversity of the YCB object dataset. We observed an overall success rate of 80\% (24 / 30), which is comparable to the success rate reported on novel objects in other recent papers applying deep learning to parallel jaw robotic grasping \cite{mahler2017dex, levine2016learning}. 

In addition to the failure modes described above, we saw two additional failure modes in real-world experiments. The first was an inability to deal with highly translucent objects like the clear lid in our object dataset. This likely results from the inability of the RealSense to provide accurate readings for such objects. The second is objects with highly nonuniform densities like the hammer in our dataset. One reason for this failure is likely that our training data pipeline does not generate objects with components of different densities. 

A video of our real-world experiments is available on the website for this paper.\footnote{https://sites.google.com/openai.com/domainrandomization-grasping/home}

\section{CONCLUSION}
We demonstrated that a grasping model trained entirely using non-realistic 
procedurally generated objects can achieve a high success rate on realistic 
objects despite no training on a realistic objects. Our grasping 
model architecture allows for efficient sampling of high-likelihood grasps at 
evaluation time, with a successful grasp being found for 96\% of objects 
in the first 20 samples. By scoring those samples, we can achieve an overall success 
rate of 92\% on realistic objects on the first attempt. We also demonstrated that models learned using our method can be transferred successfully to the real world.

Future directions that could improve the success rate of the trained models 
include scaling up to larger training sets, providing the model with feedback 
from failed grasps to influence further grasp selection, combining our grasp 
planning module with work on visual servoing for grasping, and incorporating 
additional sensor modalities like haptic feedback. 

Another exciting direction is to explore using domain randomization for 
generalization in other robotic tasks. If realistic object models are not 
needed, tasks like pick-and-place, grasping in clutter, and tool use may 
benefit from the ability to randomly generate hundreds of thousands or millions 
of 3D scenes.   

\section*{Acknowledgements}
We thank Lukas Biewald and Rocky Duan for helpful discussions, brainstorming, 
and support. The project could not have happened
without the help of Rachel Fong, Alex Ray, Jonas Schneider, Peter Welinder, and the rest of the engineering team at OpenAI.

\bibliographystyle{plain}
\bibliography{refs}


\end{document}